# Skill Analysis with Time Series Image Data

Toshiyuki MAEDA, Masanori FUJII
Faculty of Management Information
Hannan University
Matsubara, Japan

Isao HAYASHI
Faculty of Informatics
Kansai University
Takatsuki, Japan

*Abstract*—We present a skill analysis with time series image data using data mining methods, focused on table tennis. We do not use body model, but use only hi-speed movies, from which time series data are obtained and analyzed using data mining methods such as C4.5 and so on. We identify internal models for technical skills as evaluation skillfulness for the forehand stroke of table tennis, and discuss mono and meta-functional skills for improving skills.

## I. Introduction

As for human action and skill, internal structure of technical skill is layered with mono-functional skill which is generated by human intention, and meta-functional skill which is adjusted with environmental variation [T.Shiose.2004], [Y.Matsumoto.2003]. Matsumoto et al., furthermore, discuss that highly skilled workers in companies have internal models of the layered skill structures and they select an action process from internal models in compliance with situations [Y.Matsumoto.2003]. It is even difficult, however, for skilled workers to understand internal models completely by himself. They usually observe objectively their own represented actions, and achieve highly technical skills with internal models. High level skill is emerged with refinement of internal models, where some processes are smoothly collaborated such as bottom-up process from monofunctional skill into meta-functional skill, and as top-down process of arrangement of representing actions into mono and meta-functional skills [I.Hayashi.2009]. On the contrary, many researches are based on body structure model and/or skeleton structure model introduced from activity measurement or biomechatronical measurement [Y.Mochizuki.2002], [J.Kasai.1994], [M.Miyaki.1991] in the field of sport skill analysis. In our research, we assume that forehand strokes[J.Kasai.1994], [M.Miyaki.1991] of table tennis play exemplifies the typical sport action, and then identify internal models using data mining methods without body structure model nor skeleton structure model. We have researched technical skill of table tennis[T.Maeda.2009], and analyze forehand strokes from time series images. We evaluate those into 3 play levels as high/middle/low, and identify internal models using data mining methods.

## II. Analysis for Table Tennis Forehand Strokes

In researches of sports motion analysis, [H.Oka.1991] records excited active voltage of muscle fiber using on-body needle electromyography, and [J.Moribe.1991] uses a marking observation method with on-body multiple marking points, where their objects are to clarify body structure and skeleton structure. Figure 1 shows our system structure.

In our research, we provide that technical skills consist of internal models of layered structure as;

- Mono-functional skills corresponding to each body part, and
- Meta-functional skills as upper layer.

We thus identify the internal models from observed time series image data and skill evaluation with represented actions, without discussing the body structure or skeleton structure. In this paper, we focus on table tennis among various sports, and analyze table tennis skills of forehand strokes from observed time series image data and skill evaluation with represented actions.

## III. Using the Template

In our experiment, there are 15 subjects who are university male students. At first, as skill evaluation of representing action, We classify as;

- expert class: members of table tennis club at university,
- intermediate class: student who used to be members of table tennis club at junior high or high school, and
- novice class: inexperienced students.

We have recorded moving pictures of 15 subjects who are 7 expert / 3 intermediate / 5 novice university students. Each player is marked 9 points on the right arm as;

1) Acromioclaviclar joint point,

2) Acromiale point,

3) Radiale,

4) Ulna point,

5) Stylion,

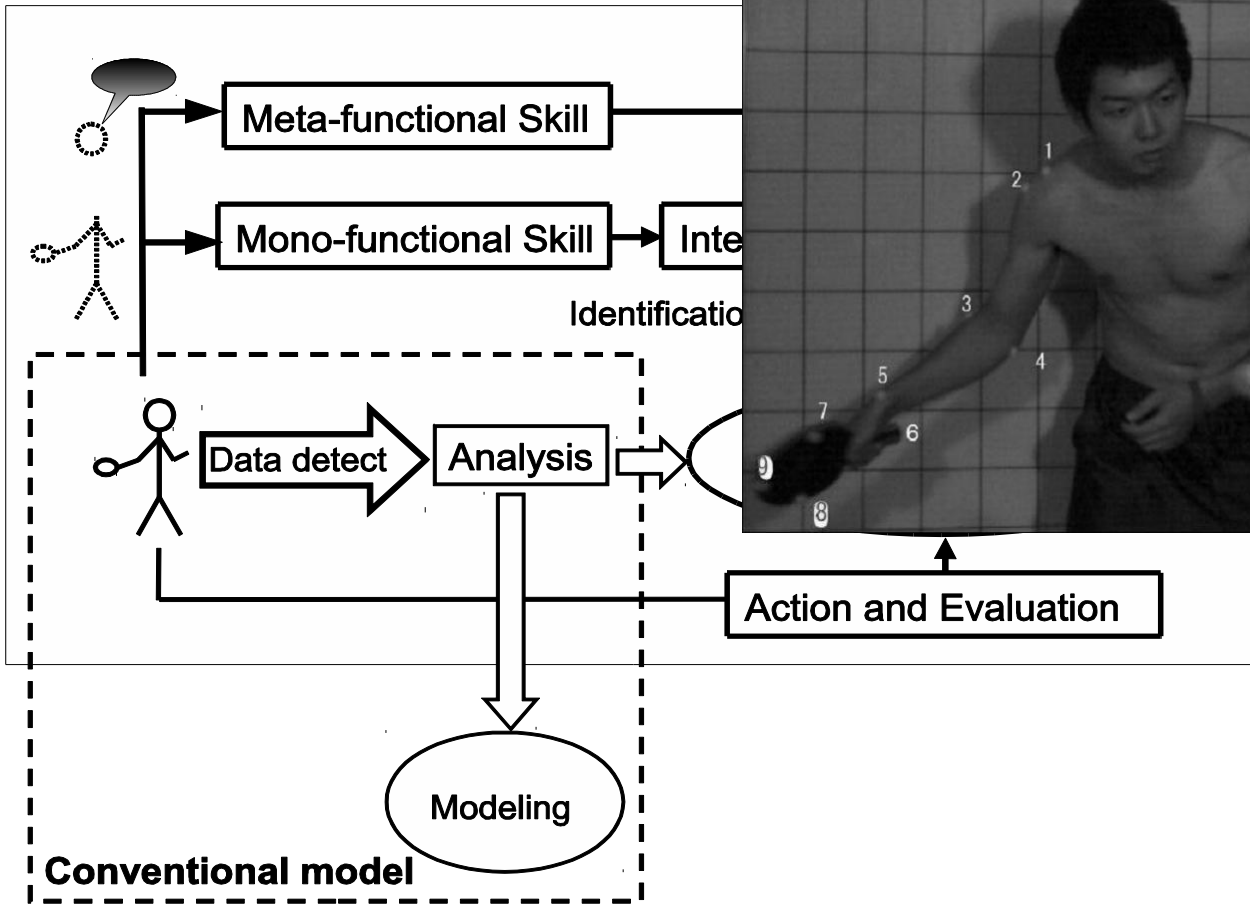

**Figure 1. Conceptional Model System**

6) Stylion ulnae,

7) Inner side of racket,

8) Outer side of racket, and

9) Top of racket.

**Figure 2.** Measurement Marking

Fig 2 shows the positions of marking sets. The ball delivery machine is installed and a subject player returns the delivered ball in a fore - cross way. We have recorded the moving traces of forehand strokes using a high-speed camcorder (resolution: 512 by 512 pixel and frame-rate: 90 fps) installed 130 cm tall and 360 cm ahead from the player. While returning the player in 10 minutes, several forehand strokes are recorded for each player (See Figure 3).

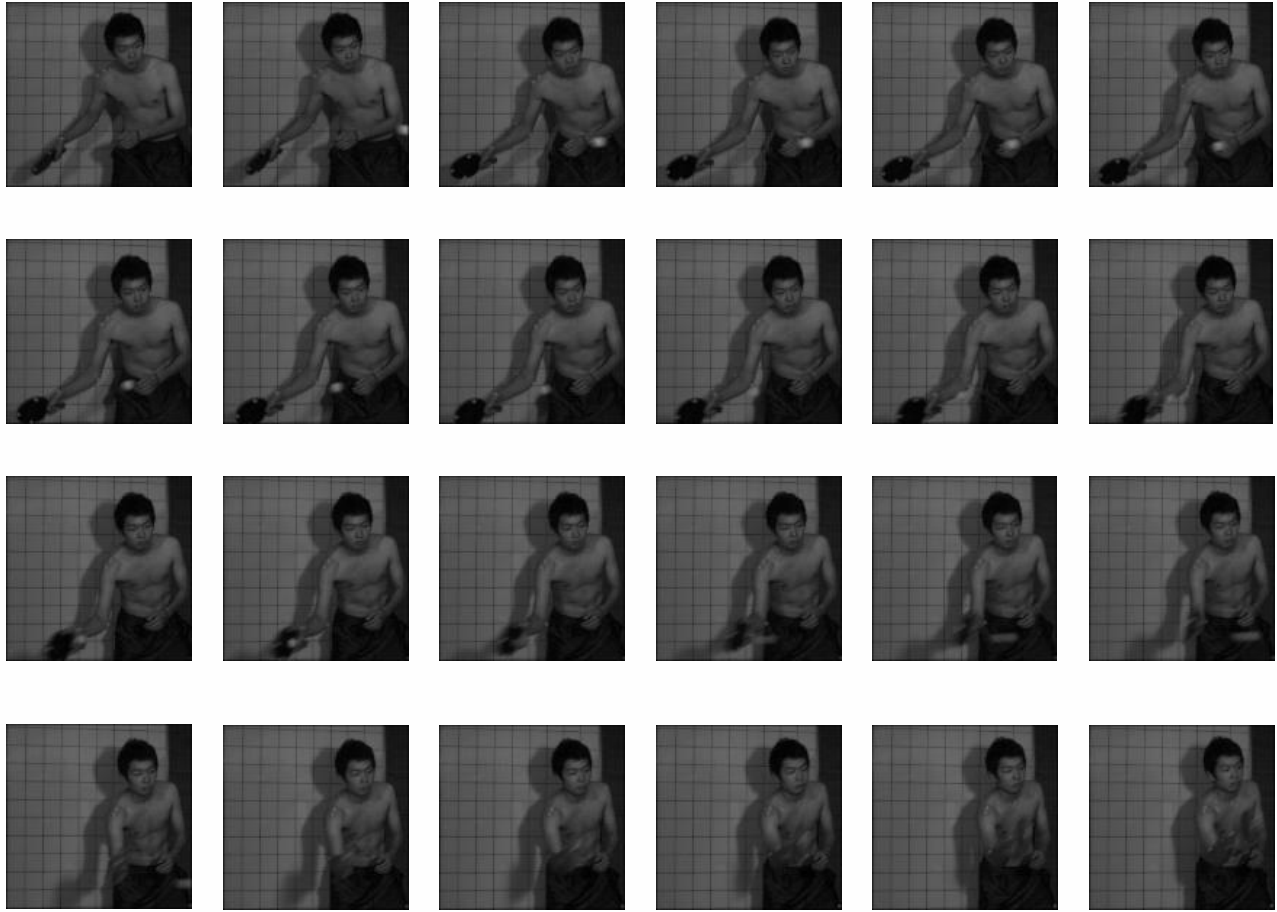

**Figure 3. Pictures of Subject.**

## IV. DATA ANALYSIS

From recorded time series images, 40 to 120 frames are retrieved from the beginning of the take-back to the ball, until the end of forehand stroke. After that, we have retrieved two dimensional axes of 9 marking points for each frame, where the starting point is set to the shoulder position of the first frame. For instance, two dimensional axes and horizontal speeds of markings for expert / intermediate / novice players are shown in Figure 4 and Figure 5.

Furthermore, Table I shows maximum and minimal values of the horizontal axes for the mark 1(M1), mark 4(M4), mark 9(M9) in Figure 4. Figure 4, Figure 5, and Table I imply as follows:

**Table 1. MIN. AND MAX. POSITION OF X-DIRECTION OF MARKINGS.**

|  | *M1* | | *M4* | | *M9* | |
|---|---|---|---|---|---|---|
|  | Min. | Max. | Min. | Max. | Min. | Max. |
| expert | -3 | 114 | -29 | 254 | -267 | 72 |
| intermediate | -10 | 116 | -25 | 236 | -218 | 577 |
| novice | -33 | 152 | -50 | 239 | -214 | 697 |

- Among expert players, there is a high correlation for marking the position of *M*1 - *M*9, where correlation coefficient is $x = 0.985; y = 0.790$. That indicates expert players have technical skills for same trajectory of swings. The trajectory, Moreover, looks less fluctuation which indicates expert players swing more smoothly.

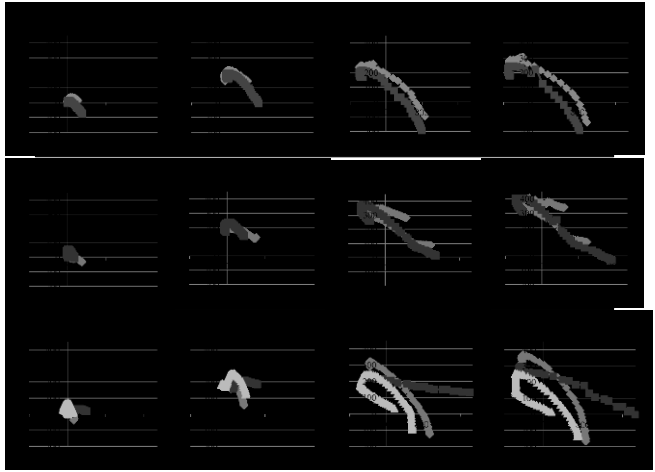

**Figure 4. Position of Markings.**

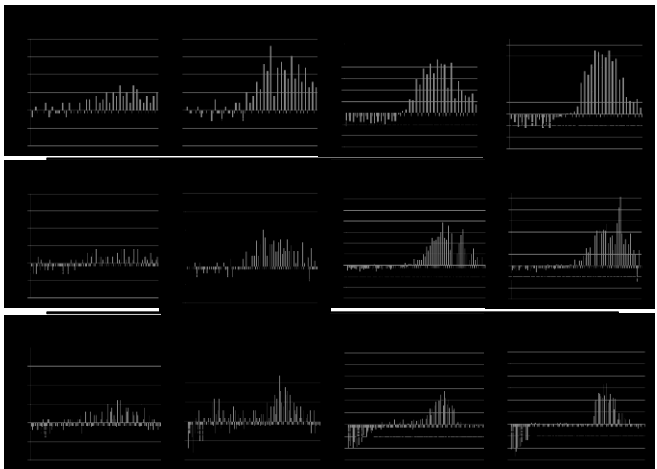

**Figure 5. Speed of Markings**

- Swing speeds of expert players show maximum at the impact of a ball-racket contact for all marking points, and that implies they have learned the technical skill of max speed impact.

- Marking positions of novice players have less correlations (correlation coefficient: $x = 0.073$; $y = -0.04$), especially position $M1$ differs much from each novice player, and that indicates novice players tend to move the shoulders. Furthermore, position axes for $M7$ and $M9$ differ for novice players and thus there is no swing trajectory. Of those, there are many variations for swings for novice players.

- All results imply that expert or intermediate players can make some categorical groups for technical skills, but there seems not to be a category for novice players because of various individual technical skills.

## V. DATA MINING FOR SKILL CLASSIFICATION

As mentioned above, one reason for decreasing the classification rate may be the existence of Middle class, as the features are not specific rather than the other two classes. The skill evaluation of represented action consists of two classes (expert / novice). Each marking position is represented two dimensional and so the observed data are reconstructed into 90-input / 2-class output. As for expert players, data of two players, which have high correlation coefficient, are used as learning data, and the rest (one player) for the evaluation.

In our experiments, technical skills for table tennis depend on trajectories rather than axes of observed making points. For applying observed data of forehand strokes of 7 subject players, we reconstruct time series data from the original data. One datum is a set of 90-tuple numbers (9markings by 2 axis(x; y) by 5 frames), and each datum is overlapped with 3 frames data (from third to fifth frame) of the next datum for presenting linkage of each datum (See Fig 6).

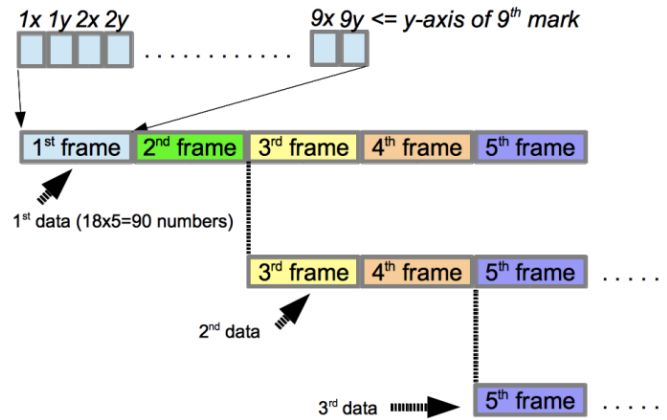

**Figure 6. Data structure from isolated pictures.**

We use an integrated data-mining environment "weka" [Weka.2012] and analyze the data by C4.5, Native Bayes Tree (NBT). Table II shows the recognition rate of modified data sets. Table III also shows the discrimination of classes for each analyzing method for evaluation data.

**Table 2. RECOGNITION RATE OF MODIFIED DATA SETS.**

|  | Recognition Rate (%) | | |
|---|---|---|---|
|  | Cross valid. | Learn. data | Eval. data |
| C4.5 | 95.6 | 98.1 | 81.2 |
| NBT | 58.9 | 58.9 | 52.8 |

**Table 3. DISCRIMINATION OF CLASSES ON C4.5.**

|  | Output class | Number of learning data | |
|---|---|---|---|
|  |  | expert | novice |
| C4.5 | expert | 40 | 0 |
|  | novice | 26 | 72 |

In those results, recognition rates of NBT for cross validation and learning data are very low. The recognition rate for evaluation data on C4.5 is quite good, though NBT makes poor results for all data. We investigate further for C4.5 analysis so that novice player classification is perfect, though some of expert players are classified into novice, which may be because of some subtle differences of swings, though they should be investigated more.

## VI. CONCLUSIONS

This paper addresses analysis and classification for internal models of technical skill with time series images of table tennis forehand stroke, and discuss experimental results. As future plans, we progress further evaluation of data, and measure more precise data and then analyze if needed.

### ACKNOWLEDGMENT

The Part of this research was supported by the Ministry of Education, Science, Sports and Culture, Japan; Grant-in-Aid for Scientific Research (C), 24520173 (2012-2014). This research was also partially conducted under a sponsorship by Grant of Institute of Industrial and Economic Research, Hannan University. This research was collaborated with members of the table tennis club at Hannan University, students at Hannan University and Kansai University. The authors greatly appreciate those.